\documentclass[sigconf]{aamas} 

\usepackage{balance} 

\usepackage[ruled,vlined]{algorithm2e}

\usepackage{algorithmic}
\usepackage{mathtools}
\usepackage{diagbox}
\usepackage{bm}
\usepackage{subcaption}
\usepackage{todonotes}
\DeclarePairedDelimiter\ceil{\lceil}{\rceil}
\DeclarePairedDelimiter\floor{\lfloor}{\rfloor}

\newtheorem{theorem}{Theorem}

\setcopyright{ifaamas}
\acmConference[AAMAS '21]{Proc.\@ of the 20th International Conference on Autonomous Agents and Multiagent Systems (AAMAS 2021)}{May 3--7, 2021}{Online}{U.~Endriss, A.~Now\'{e}, F.~Dignum, A.~Lomuscio (eds.)}
\copyrightyear{2021}
\acmYear{2021}
\acmDOI{}
\acmPrice{}
\acmISBN{}

\title[AAMAS-2021 Formatting Instructions]{Sequential Ski Rental Problem}

\author{Anant Shah}
\affiliation{
  \institution{Department of Electrical Engineering}
  \city{Indian Institute of Technology Madras}
  }
\email{anantshah200@gmail.com}

\author{Arun Rajkumar}
\affiliation{
  \institution{Department of Computer Science and Engineering}
  \city{Indian Institute of Technology Madras}
  }
\email{arunr@cse.iitm.ac.in}

\begin{abstract}
The classical ‘buy or rent’ ski-rental problem was recently considered in the setting where multiple experts (such as Machine learning algorithms) advice on the length of the ski season. Here, robust algorithms were developed with improved theoretical performance over adversarial scenarios where such expert predictions were unavailable. We consider a variant of this problem which we call the ‘sequential ski-rental’  problem. Here, a sequence of ski-rental problems has to be solved in an online fashion where both the buy cost and the length of ski season are unknown to the learner. The learner has access to two sets of experts, one set who advise on the true cost of buying the ski and another set who advise on the length of the ski season. Under certain stochastic assumptions on the experts who predict the buy costs, we develop online algorithms and prove regret bounds for the same. Our experimental evaluations confirm our theoretical results.
\end{abstract}


\newcommand{\BibTeX}{\rm B\kern-.05em{\sc i\kern-.025em b}\kern-.08em\TeX}

\begin{document}


\pagestyle{fancy}
\fancyhead{}


\maketitle 


\section{Introduction}

The classic buy or rent ski-rental problem is the following: Given a buy cost $b$ for the ski gear, one needs to make a decision on each day whether to rent the ski for a cost of $1$ unit or end the sequence of decisions by buying the ski for a cost of $b$ units. The challenging part of the problem is that one does not have apriori knowledge of how long the ski season would last. The ski-rental problem has found several practical applications including TCP acknowledgement, time scheduling, etc. In the basic version of the problem, a strategy achieving a competitive ratio of $2$ is to rent for $b$ days (if the ski season lasts) and buy thereafter. A recent line of work tries to consider the same problem when certain expert advice (potentially machine learning predictions) is available about the length of the ski season. In this scenario, it was shown recently that one can do better when the predictions are within reasonable deviations from the truth. While current work in this area( \cite{PSK-18}, \cite{GP-19}) propose algorithms that require a truthful environment parameter (buy cost), our main contribution is in the case the environment parameter is predicted by a set of experts.

In this work we introduce a variant of the problem which we term the \emph{sequential ski rental problem}. Here, the ski season proceeds in a sequence of rounds and in each round, the player has to come up with a strategy to buy/rent a \emph{perishable} item every day of that season. The item if bought lasts only for that season and becomes unavailable the following seasons. In addition to the unavailability of the length of the season, the player also has to make an \emph{early} rent/buy decision \emph{before} the buy cost is revealed. The problem thus has two levels of uncertainty, in the ski-length and the buy costs respectively.

 At first glance, such a problem might seem hopeless to solve as there is no input parameter for the algorithm to work with (i.e., neither the buying price nor the length of the ski season). To tackle this, we consider two types of advice as being available to the player - one on the length of the ski season and other on the buy cost. Both of these could come from potential machine learning prediction models or from certain noisy inputs. In this novel setting, we develop online learning algorithms which have good performance with respect to the worst case bounds. Our algorithm relies on a combination of the stochastic Hedge algorithm and the algorithm for ski-rental problem with expert advice. Our main contributions are as follows
\begin{itemize}
\item We introduce the Sequential ski rental problem with two sets of expert advice
\item We develop novel algorithms with performance guarantees
\item We demonstrate the efficacy of the algorithm on experiments.
\end{itemize}

Our work can be applied to various real world settings an example of which is the following. Consider a customer on an online retail site, looking to purchase different products in a sequential fashion. For each item she desires, she needs to make a decision on whether to buy or rent the product. The way she decides her strategy is based on a set of experts who are given as input a buy cost and output a rent-buy strategy. The online retail site has a fixed purchase price for each product. However, the customer will need to pay an additional overhead cost which is unknown apriori. A main source of this overhead cost could be delivery charges (or more abstract costs involving time delays in delivery etc). This exact overhead cost is not clear to the customer at the time of deciding her strategy and she only has access to an estimate of this cost. The problem then is how well the customer performs when she is given the estimated cost at the time of deciding her strategy as compared to what she would have done had she been given the exact overall cost to decide her strategy. Clearly since the customer can purchase different items, the buy cost varies.

\section{Related Work}

Our work is closely related to \cite{GP-19} and \cite{PSK-18}. We utilize the concepts of robustness and consistency introduced in \cite{LV-18}, who utilized predictions for the online caching problem . \cite{GP-19} consider the setting in which multiple experts provide a prediction to the ski-rental problem. They obtain tight bounds for their settings thereby showing the optimal robustness and consistency ratios in the multi-expert setting. However their algorithm depends on the fact that the error in the buy cost is zero, in contrast to our setting where the buy-cost given to the experts may not be true as they are forced to make an early decision. 

For the ski-rental problem, \cite{PSK-18} utilize predictions from a single expert to improve on the lower bound of $\frac{e}{e-1}$, shown to be optimal by \cite{KMMO-94} . \cite{PSK-18} design a robust, consistent deterministic and randomized algorithm which performs optimally for a zero error prediction, while at the same time matches the optimal lower bound in the case the prediction has infinite error. As was the case in \cite{GP-19}, their algorithm depends on the fact that the buy cost is without error. Even for a small perturbation in the buy-cost, the expected algorithmic cost suffered is different. As a part of our online learning subroutine, we propose an algorithm which is robust to small perturbations in the environment parameters, the environment parameter being the buy-cost for the ski-rental problem. 

Recent work in online algorithms with machine learned predictions extend to numerous applications such as online scheduling with machine learned advice (\cite{STBS-20}, \cite{rohatgi2020near}, \cite{JPS-20}) or to the reserve price revenue optimization problem which utilizes predictions of the bids by bounding the gap between the expected bid and the revenue in terms of the average loss of the predictor (\cite{MS-17}). 

Our work tackles the problem of uncertainty by utilizing the online learning model. Specifically, we have two sets of experts where the loss value of one set comes from a stochastic distribution. Well studied models to tackle this uncertainty include that of robust optimization (\cite{KY-13}) which gives good guarantees for potential realizations of the inputs. Another model is that of stochastic optimization (\cite{BS-12}, \cite{MGZ-12}, \cite{MNS-12}) where the input comes from a known distribution. Our work utilizes the multiplicative weight update (\cite{LW-89}), also known as the Constant Hedge algorithm, as our learning algorithm for one set of experts. We utilize the Decreasing Hedge algorithm (\cite{ABG-02}) to update the weights of another set of experts whose losses come from a stochastic distribution. Considering this easier setting for learning, (\cite{HR-15}, \cite{NYG-07}, \cite{PGT-14}, \cite{AGA-14}, \cite{WPT-16}, \cite{STPW-14}, \cite{JS-19}) tackle this by designing algorithms that rely on data dependent tuning of the learning rate or better strategies and give theoretical results on the regret bounds in these settings.

An interpretation of our work is to the classic online learning set-up given in a survey by \cite{Sha-11}. Instead of each hypothesis receiving the true environment parameter, based on which they make a recommendation to the learner, they receive an unbiased sample with some noise. Something similar has been studied in the area of differential privacy (\cite{DNP-10}) , where certain privacy guarantees are shown if one of the loss vectors in the loss sequence changes. Our model essentially boils down to the scenario where the initial sequence of loss vectors are different but over time they converge to the true loss sequence.



\section{Preliminaries}

We consider the online learning model to the ski-rental problem. 

\textbf{Ski-Rental Problem} : In the ski-rental problem, an example of a large class rent-or-buy problem, a skier wants to ski and needs to make a decision whether to buy the skis for a cost of $b$ (non-negative integer) units or rent the skis for a cost of 1 unit per day. The length of the ski season $x$ (non-negative integer) is \emph{not known}. Trivially, if the ski season lasted more than $b$ days and the skier knew this beforehand, she would buy them at day 1, otherwise she would rent them for all days. The minimum cost that can be suffered is $OPT=\min\{b,x\}$. For this problem, the best a deterministic algorithm can do is obtain a competitive ratio of $2$, while \cite{KMMO-94} designed a randomized algorithm which obtains a competitive ratio of $\frac{e}{e-1}$ which is optimal.

\textbf{Robustness and Consistency} : We utilize the notions of robustness and consistency which are defined when online algorithms utilize machine learned predictions. In online algorithms, the ratio of the algorithmic cost($ALG$) to the optimal offline cost($OPT$) is defined as the competitive ratio. While utilizing predictions, such a ratio would be a function of the accuracy $\eta$ of the predictor. Note that the algorithm has no knowledge about the quality of the predictor. An algorithm is $\alpha$- robust if $\frac{ALG(\eta)}{OPT} \leq \alpha$ for all $\eta$ and is $\beta$-consistent if $\frac{ALG(0)}{OPT} \leq \beta$. The goal is to use the predictions in such a way that if the predictions are true, the algorithm performs close to the offline optimal and even if the prediction is very bad, it performs close to the original online setting without any predictions. 

\textbf{Hedge Algorithm for the expert advice problem} : In the classic learning from expert advice setting, also known as decision-theoretic online learning (\cite{FS-97}), the learner maintains a set of weights over the experts and updates these weights according to the losses suffered by the experts. These losses could potentially be chosen in an adversarial manner. Specifically, the learner has weights $\bm{\alpha^{t}} = (\alpha^{t}_{i})_{1 \leq i \leq M}$ over $M$ experts at time $t$. The environment chooses a bounded, potentially adversarial, loss vector $\bm{l^{t}}$ over these experts at time $t$. The loss suffered by the learner is $(\bm{\alpha^{t})}^{T}\bm{l^{t}}$. The goal of the learner is to compete against the expert with the minimum cumulative loss, i.e to minimize the \emph{regret} which is defined as

$$R_{T} = \sum_{t=1}^{T}(\bm{\alpha^{t})}^{T}\bm{l^{t}} - \min_{i \in [M]}\sum_{t=1}^{T}l^{t}_{i}$$

where $T$ denotes the number of instances for which this game is played between the learner and the environment. We say that an algorithm is a no-regret learning algorithm if the regret is sub-linear in $T$ .The multiplicative weights algorithm(\cite{LW-89}) updates the weights \emph{optimally} as 

$$\alpha^{t}_{i} = \frac{e^{-\eta_{t}\sum_{t'=1}^{t'=t-1}l^{t'}_{i}}}{\sum_{i=1}^{M}e^{-\eta_{t}\sum_{t'=1}^{t'=t-1}l^{t'}_{i}}}$$

where $\eta_{t}$ is the learning rate. The learning rate could be constant or dependent on $t$. The Decrease Hedge algorithm(\cite{ABG-02}) has a learning rate $\eta_{t} \propto 1/\sqrt{t}$ while the Constant Hedge algorithm(\cite{LW-89}), given a $T \geq 1$, has a learning rate $\eta_{t} \propto 1/\sqrt{T}$. The standard regret bound for the hedge algorithm (eg : \cite{CZ-10}) is sub-linear in $T$. 

We look at the learning from expert advice problem from a different lens. At each stage $t$, there exists an environment parameter $\mathbf{e}^{t}$ based on which the experts give their recommendation. The recommendation can be the output of some machine learned model which each of the $M$ experts have, denoted by $(h_{i}(\mathbf{e}^{t}))_{1 \leq i \leq M}$. A more general view of the problem we tackle in this paper is what happens in the case the experts have access to some estimate $\mathbf{\hat{e}^{t}} \neq \mathbf{e}^{t}$ such that $\mathbf{E[\mathbf{\hat{e}^{t}}]} = \mathbf{e}^{t}$. The constraint we use on this setting is that the variance of the estimator $\hat{\mathbf{e}^{t}}$ becomes small as $t$ becomes large. 

\textbf{Stochastic Setting} : When the losses are realizations of some unknown random process, we consider it as the stochastic setting. Our work considers the standard i.i.d case where the loss vectors are $\bm{l^{1}},\bm{l^{2}},\dots,\bm{l^{t}}$ which are i.i.d. In general there need not be independence across experts. We define the sub-optimality gap as $\Delta = \min_{i \neq i^{*}} \mathbf{E}[l^{t}_{i}-l^{t}_{i^{*}}]$ where $i^{*} = argmin_{i}\mathbf{E}[l^{t}_{i}]$. A natural extension is whether the hedge algorithm obtains a better regret guarantee in the nicer stochastic setting. \cite{JS-19} show that for the Decreasing Hedge algorithm, a better regret bound in terms of the sub-optimality gap $\Delta$ can be obtained while also showing that for the Constant Hedge algorithm the $\sqrt{T\log M}$ regret bound is the best possible. A part of our setting is inspired by \cite{JS-19} as we have a set of experts whose loss comes from a stochastic distribution.

\section{Problem Setting}

In the ski-rental setting with predictions, the rental costs are 1-unit per day, $b$ is the buy cost, $x$ is the true number of ski days which is unknown to the algorithm and $y$ is the predicted number of ski days. We use $\eta = |y-x|$ to denote the prediction error. No assumptions are made on how the length of the ski season is predicted. The optimal strategy in hindsight will give us an optimal cost of $OPT = \min \{b,x\}$.

\subsection{Sequential Ski Rental Setup}

Our learning model has two sets of experts, one set predicting the environment parameters i.e the buy cost of the skis. Let there be $m$ such experts. We will call them as buy-experts from now on. The other set of experts are those that are giving advice to the learner on what strategy to follow based on their prediction of the number of ski-days. They utilize the prediction of the buy-experts to decide their strategy. Let these experts be $n$ in number. We will call them as ski-experts from now on.   

We are running multiple ski-rental instances over the time horizon $T$. For each $t \in [T]$, we denote the ground truth buy cost as $b^{t}$ and the ground truth ski-days as $x^{t}$. The ski experts only make a prediction on $x^{t}$ denoted by $\bm{y}^{t} = (y^{t}_{j})_{1\leq j \leq n}$, which is a vector of non-negative integers, and suggest a strategy to the learner based on the predicted value of the buy experts, which we denote by $b^{t}_{s}$. The way a ski-expert utilizes its prediction $y^{t}_{j}$ to suggest a strategy is based on a randomized algorithm which at a high level suggests to buy late if $y^{t}_{j} < b^{t}_{s}$ or suggests to buy early
if $y^{t}_{j} \geq b^{t}_{s}$. 

\subsection{Buy Expert Predictions}

We assume that the buy costs over rounds are integers in the range $[2,B]$ where $B$ is finite. In our model we consider that each buy cost prediction comes from a stochastic distribution with mean equal to the ground truth buy cost and a certain variance which corresponds to the quality of that buy expert. Let best buy expert be $i^{*}$ and her variance be $\gamma_{min}$. We define the sub-optimality gap for buy experts $i \neq i^{*}$ in terms of the variance as $\Delta_{i} = \gamma_{i} - \gamma_{min}$ where $\gamma_{i}$ corresponds to the variance of the $i^{th}$ buy expert. Let the vector of these predictions be $\bm{a^{t}} = (a^{t}_{i})_{1 \leq i \leq m}$. In our setting,

$$\bm{a}^{t} = b^{t}\vv{1} + \bm{\epsilon^{t}_{b}}$$

where $\bm{\epsilon_{b}^{t}}$ has zero mean and its covariance matrix is a diagonal matrix due the independence across experts. We assume that $\bm{\epsilon^{t}_{b}} \in [-1,1]^{m} $. The algorithm maintains a weight vector $\bm{\alpha}^{t}$ over the buy experts corresponding to it's confidence over that particular expert. We use the Decreasing Hedge algorithm to update these weights, where the loss function is the squared error loss

$$\alpha^{t}_{i} = \frac{e^{-\eta_{t}\sum_{t'=1}^{t'=t-1}(a^{t'}_{i}-b^{t'})^{2}}}{\sum_{i=1}^{m}e^{-\eta_{t}\sum_{t'=1}^{t'=t-1}(a^{t'}_{i}-b^{t'})^{2}}}$$

The ski-experts are given the buy cost prediction $b^{t}_{s}$ on which they base their strategy

$$b^{t}_{s} = (\bm{a^{t}})^{T}\bm{\alpha^{t}}$$


\subsection{Ski Expert Predictions}

The ski-experts suggest a strategy to the learner and suffer some loss for the same. They make a prediction on the number of ski days and then suggest a strategy to the learner utilizing the predicted buy cost $b^{t}_{s}$. While we do make an assumption on the prediction distribution, specifically that it is unbiased, of the buy experts, we make no assumption on how the ski expert predictions are obtained. The learner maintains a set of weights $\bm{\beta^{t}}$ over these experts which are updated using the Constant Hedge algorithm. The loss suffered by the $j^{th}$ ski-expert at time $t$ is denoted by $l^{t}_{j}(b^{t},x^{t},b^{t}_{s},y^{t}_{j})$. We denote the loss vector suffered by these experts as $\bm{l}^{t}(b^{t},x^{t},b^{t}_{s},\bm{y}^{t})$.  

\subsection{Regret}

Our learning setup is as follows. At each $t \in [T]$ the learner chooses a strategy recommended by a ski-expert by sampling from the distribution $\bm{\beta}^{t}$ over these experts and suffers an expected loss. The ski-experts are in turn basing their strategy on a prediction of the buy cost from the buy-experts. At a time $t$, the loss suffered by the $j^{th}$ ski-expert depends on the ground truth values and the predictions it receives and is hence denoted by $l^{t}_{j}(b^{t},x^{t},b^{t}_{s},y^{t}_{j})$. 

The goal of the learner is to compete against the best expert in the case the experts are given the true environment parameters, that being the buy cost in our case. This lends to the expected regret definition which we wish to minimize



$$R_{T} = \sum_{t=1}^{T}(\bm{\beta^{t}})^{T}\bm{l}^{t}(b^{t},x^{t},b^{t}_{s},\bm{y^{t}}) - \min_{j}\sum_{t=1}^{T}l^{t}_{j}(b^{t},x^{t},b^{t},y^{t}_{j})$$

\begin{table}[]
    \centering
    \begin{tabular}{c|c}
    \hline
        Symbol & Description  \\
        \hline
        $x^{t}$ & True number of ski-days at the $t^{th}$ time instant. \\
        \hline
        $y^{t}_{j}$ & $j^{th}$ expert ski-day prediction at time $t$ \\
        \hline
        $b^{t}$ & True buy cost at time $t$ \\
        \hline
        $a^{t}_{i}$ & $i^{th}$ expert buy cost prediction at time $t$\\
        \hline
        $b^{t}_{s}$ & Weighted sum estimate of the buy cost
    \end{tabular}
    \caption{Notation}
    \label{tab:notation}
\end{table}

\section{Algorithm}

We introduce the subroutine used by the ski-experts to compute a strategy, which we call the \emph{CostRobust Randomized Algorithm} and propose the \emph{Sequential Ski Rental} algorithm which the learner utilizes when she has access to two sets of experts.

\subsection{CostRobust Randomized Algorithm}

Below we present an algorithm which obtains robust and consistent results for solving the ski-rental problem. The crucial difference between our algorithm and \cite{PSK-18} is that our algorithm is robust to small variations in the buy cost i.e if we obtain a noisy sample of the buy cost, our algorithm suffers a cost which is same as the cost suffered if the true value were given. Our algorithm obtains similar consistency and robustness guarantees while considering the competitive ratio.

\begin{algorithm}
\DontPrintSemicolon
\SetAlgoLined

\SetKwFunction{Floss}{SkiRentStrategy}
\SetKwProg{Fn}{Function}{:}{}

\Fn{\Floss{$b,y$}}{
\eIf{$y \geq nint(b)$}{
$k \gets \floor*{\lambda b}$ \;
$q_{i} \gets (1-\frac{\lambda}{k})^{k-i}.\frac{\lambda}{k(1-(1-\lambda/k)^{k})} \forall 1 \leq i \leq k$ \;
$d \sim \bm{q}$ // Sample the buy day based on the distribution above \;
return $d$ \;
}{

$l \gets \ceil*{ b/\lambda}$ \;
$r_{i} \gets (1-\frac{1}{\lambda l})^{l-i}.\frac{1}{l \lambda(1-(1-\frac{1}{\lambda l})^{l})} \forall 1 \leq i \leq l$ \;
$d \sim \bm{r}$ // Sample the buy day based on the distribution above \;
return $d$ \;
}
}

 \caption{CostRobust Randomized Algorithm}

\end{algorithm}

Let $\lambda \in (1/b,1)$ be a hyper-parameter. For a chosen $\lambda$ the algorithm samples a buy day from two different probability distributions depending on the prediction and the input buy cost. The algorithm outputs a buy day strategy for the prediction on the number of ski days $y$ (a non-negative integer) and the input it gets on the cost $b$. Note that $nint(.)$ is the nearest integer function where half integers are always rounded to even numbers.

We say that the algorithm is $\epsilon$ robust in terms of the buy cost if $\epsilon$ is the maximum possible value such that for a input non-negative integer $b$, if the buy cost prediction $b_{s}$ lies in the range $(b-\epsilon,b+\epsilon)$, the incurred cost is equal to the cost in the case the true value $b$ were given. Note that $nint(b) = b$ for the true buy cost. 


\begin{theorem}
The CostRobust randomized algorithm is $\epsilon$ robust in terms of the buy cost where $\epsilon$ is 

$$\epsilon = \min \left (\frac{1}{\lambda} \min (\{\lambda b\},1-\{\lambda b\}), \lambda \min \left (\left \{\frac{b}{\lambda} \right \},1- \left \{\frac{b}{\lambda} \right \} \right )  \right )$$

where $\{x\}$ denotes the fractional part of $x$.
\end{theorem}

\begin{proof}
Consider the case when $y \geq b$. In this case $k = \floor*{\lambda b}$. A predicted buy cost $b_{s}$ is given where $b-\epsilon < b_{s} < b + \epsilon$, and hence the condition we get on $\epsilon$ so that $\floor*{\lambda b} = \floor*{\lambda b_{s}}$ is 

$$\epsilon = \frac{1}{\lambda} \min (\{\lambda b\},1-\{\lambda b\})$$

Similarly, in the case $y < b$, performing a similar analysis where we require the $l$ values to be equal gives us the condition

$$\epsilon = \lambda \min \left (\left \{\frac{b}{\lambda} \right \},1- \left \{\frac{b}{\lambda} \right \} \right )  $$

Hence the result follows.

\end{proof}

We now show consistency and robustness guarantees, in terms of the competitive ratio, of our proposed algorithm in the case it receives the true prediction of the buy cost. Our analysis is similar to \cite{PSK-18}, who calculate the expected loss of the algorithm based on the relative values of $y,b,x$ and the distribution defined.

\begin{theorem}
The CostRobust randomized ski-rental algorithm yields a competitive ratio of at most $\min \{\frac{1+1/\floor*{\lambda b}}{1-e^{-\lambda}},(\frac{\lambda}{1-e^{-\lambda}})(1+\frac{\eta}{OPT})\}$ where $\lambda$ is a hyper-parameter chosen from the set $\lambda \in (1/b,1]$. The CostRobust Randomized algorithm is $\frac{1+1/\floor*{\lambda b}}{1-e^{-\lambda}}$-robust and $(\frac{\lambda}{1-e^{-\lambda}})$-consistent. 

\end{theorem}

\begin{proof}
We consider different cases depending on the values of $y,b,x,k$. Note that as $b$ is a non-negative integer, $nint(b) = b$. 
 
 \begin{itemize}
     \item $y \geq b$, $x \geq k$ Based on the algorithm $k=\floor*{\lambda b}$. 
     
     \begin{equation*}
         \begin{split}
             \mathbf{E}[ALG] & = \sum_{i=1}^{k}(b+i-1)q_{i} \leq b - \frac{k}{\lambda} + \frac{k}{1-(1-\frac{\lambda}{k})^{k}} \\
             & \leq b - \frac{k}{\lambda} + \frac{k}{1-e^{-\lambda}} 
             \leq b + b \left (\frac{\lambda}{1-e^{-\lambda}}-1 \right) \\
             & \leq \left (\frac{\lambda}{1-e^{-\lambda}} \right)(OPT+\eta)
         \end{split}
     \end{equation*}
     
     where the second to last inequality from the fact $b > \frac{k}{\lambda}$ and the last inequality from the fact $y>b$.
     
     \item $y \geq b$, $x < k$. For this ordering of the variables $OPT=x$. The algorithm suffers a loss of $ALG = b+i-1$ if it buys at the beginning of day $i \leq x$. Thus ,
     
     \begin{equation*}
         \begin{split}
             \mathbf{E}[ALG] &= \sum_{i=1}^{x}(b+i-1)q_{i} + \sum_{i=x+1}^{k}x q_{i} \\
             & = \frac{x}{1-(1-\frac{\lambda}{k})^{k}} \left (1+\left (1-\left (1-\frac{\lambda}{k} \right)^{x} \right) \left (1-\frac{\lambda}{k} \right )^{k-x}\frac{(b-\frac{k}{\lambda})}{x} \right ) \\
             & \leq \left (\frac{1}{1-e^{-\lambda}} \left (\frac{\lambda b}{\floor*{\lambda b}}\right ) \right)OPT \\
             & \leq \left (\frac{1+1/\floor*{\lambda b}}{1-e^{-\lambda}} \right)OPT
         \end{split}
     \end{equation*}
     
     where the second to last inequality follows from the fact that $(1-\frac{\lambda}{k})^{x} > 1 - \frac{\lambda x }{k}$. To show consistency, we have the inequality
     
     \begin{equation*}
         \begin{split}
             \mathbf{E}[ALG] & \leq b - \frac{k}{\lambda} + \frac{x}{1-e^{-\lambda}} \leq \frac{\{\lambda b\}}{\lambda} + \frac{x}{1-e^{-\lambda}} \\
             & \leq \frac{\{\lambda b\}+x}{1-e^{-\lambda}} \leq \frac{\lambda b}{1-e^{-\lambda}} \\
             & \leq   \left (\frac{\lambda}{1-e^{-\lambda}} \right )(OPT+\eta)
         \end{split}
     \end{equation*}
     
     where the third inequality comes from the fact that $\lambda \geq 1-e^{-\lambda}$ for all $\lambda \in [0,1]$ and the fourth inequality comes from the fact that $x < k$.
     
     \item $y < b$, $x < l$. Based on the algorithm, $l= \ceil*{\frac{b}{\lambda}}$. The algorithm suffers a loss of $ALG = b+i-1$ if it buys at the beginning of day $i \leq x$. Thus ,
\begin{equation*}
    \begin{split}
     \mathbf{E}[ALG] &= \sum_{i=1}^{x}(b+i-1)r_{i} + \sum_{i=x+1}^{l}x r_{i} \leq (b-\lambda l) + \frac{x}{1-(1-\frac{1}{\lambda l})^{l}} \\
     & \leq \frac{x}{1-e^{-\frac{1}{\lambda}}} \leq \left (\frac{1}{1-e^{-1/\lambda}} \right )(OPT+\eta) \\
     & \leq \left (\frac{\lambda}{1-e^{-\lambda}} \right )(OPT+\eta)
     \end{split}
\end{equation*}

     \item $y < b$, $x \geq l$ . Here $OPT=b$. The expected cost incurred is
     
     \begin{equation*}
         \begin{split}
             \mathbf{E}[ALG] & = \sum_{i=1}^{l}(b+i-1)r_{i} \leq b - \lambda l + \frac{l}{1-(1-\frac{1}{\lambda l})^{l}} \\
             & \leq b + l \left (\frac{1}{1-e^{-1/\lambda}} - \lambda \right) \leq b + l \left (\frac{\lambda e^{-\lambda}}{1-e^{-\lambda}} \right) \\
             & \leq \left (\frac{1+\lambda e^{-\lambda}/b}{1-e^{-\lambda}} \right)OPT < \frac{(1+1/\floor*{\lambda b})}{1-e^{-\lambda}}OPT
         \end{split}
     \end{equation*}
   
   which shows some sense of robustness. To show consistency, we can write the equations as
   
   \begin{equation*}
       \begin{split}
           \mathbf{E}[ALG] & \leq \frac{l}{1-e^{-1/\lambda}} = \frac{1}{1-e^{-1/\lambda}}(b+l-b) \\
           & \leq \frac{1}{1-e^{-1/\lambda}}(OPT+\eta) \leq \left (\frac{\lambda}{1-e^{-\lambda}} \right)(OPT+\eta)
       \end{split}
   \end{equation*}
     
 \end{itemize}
 
Hence the result follows.
 
\end{proof}

This result provides a trade-off between the consistency and robustness ratios. Setting $\lambda=1$ gives us a guarantee that even if the prediction has a very large error($\eta \to \infty$), our competitive ratio is bounded by $\frac{e}{e-1}(1+1/b)$ which is close to the best case theoretical bound without predictions. However if we are very confident in the prediction, i.e confident that $\eta=0$, then we can set $\lambda$ to be very small and get a guarantee of performing close to the offline optimal.

\subsection{Sequential Ski Rental Algorithm}

We utilize the CostRobust algorithm as a subroutine for each ski-expert. Each of the $n$ ski experts are running the algorithm for each instance of the ski-rental problem to determine a strategy using the predicted buy cost and its own prediction on the number of ski days. The loss is calculated with respect to the best hindsight strategy and hence is always positive. We normalize the competitive ratio by an additive factor so that if an expert predicts correctly, she obtains 0 loss. 

\begin{algorithm}
\DontPrintSemicolon
\SetAlgoLined

\SetKwFunction{Floss}{loss}
\SetKwProg{Fn}{Function}{:}{}

\Fn{\Floss{$b^{t},x^{t},b^{t}_{s},y^{t}_{j}$}}{

$d \gets $ SkiRentStrategy($b^{t}_{s},y^{t}_{j}$)

\eIf{$x^{t} \geq b^{t}$}{
$OPT = b^{t}$ \;}{$OPT = x^{t}$ \;}

Based on the strategy suggested by the expert in Algorithm 1, $d$ is the buy-day.

\eIf{$x^{t} \geq d$}{
$ALG = b^{t} + d - 1$ \;}{$ALG = x^{t}$ \;}

$l^{t}_{j} = \frac{ALG-OPT}{OPT}$ // Loss suffered by the $j^{th}$ ski expert at the $t^{th}$ iteration \;

}

 \caption{Loss calculation for each expert}

\end{algorithm}

At each instance of the \emph{Sequential Ski Rental Algorithm}, the input to the \emph{CostRobust} subroutine is an estimate $b^{t}_{s} \neq b^{t}$. The statement below shows that even if the ski-days prediction of a ski-expert has very large error, it suffers a finite loss when it uses an estimate of the buy cost.

\begin{theorem}
The loss suffered by each ski expert is bounded for every round $t \in [T]$ when the buy predictions are coming from bounded random variables. 
\end{theorem}


\begin{algorithm}

\SetAlgoLined 

\textbf{Input:}  \newline
$\lambda$ : Hyperparamter \;


\textbf{Initialization:} 

$\bm{w_{\beta}^{1}} \gets (1,1,\dots,1)$ : Weights corresponding to the ski experts \;
$\bm{w_{\alpha}^{1}} \gets (1,1,\dots,1)$ : Weights corresponding to the buy experts \;

\For{$t\gets1$ \KwTo $T$}{
    input $\bm{a^{t}}$ : Buy expert predictions \;
    $\bm{\alpha^{t}} = \frac{\bm{w_{\alpha}^{t}}}{\sum_{i}(w_{\alpha}^{t})_{i}}$ : Probability distribution over the buy experts \;
    $b^{t}_{s} \gets \bm{a^{t}} \cdot \bm{\alpha^{t}}$ : The weighted buy cost prediction to be provided to the ski experts \;
    input $\bm{y^{t}}$ : Ski-expert predictions \;
    $\bm{\beta^{t}} \gets \frac{\bm{w_{\beta}^{t}}}{\sum_{j}(w_{\beta}^{t})_{j}}$ : Probability distribution over the ski experts \;
    $\bm{l^{t}} \gets loss(b^{t},x^{t},b^{t}_{s},\bm{y^{t}})$ : Loss vector where $j^{th}$ element corresponds to the loss of the $j^{th}$ ski expert \;
    
    $\bm{w_{\beta}^{t+1}} \gets \bm{w_{\beta}^{t}} e^{-\epsilon_{s} \bm{l^{t}}}$ : Update the weights according to the Constant hedge algorithm \;
    $\bm{w_{\alpha}^{t+1}} \gets \bm{w_{\alpha}^{t}} e^{-\epsilon_{b} (\bm{a^{t}}-b^{t})^{2}}$ : Update the weights according to the Decreasing Hedge algorithm
    }
    
\caption{Sequential Ski Rental Algorithm}

\end{algorithm}

We will denote this bound as $B$. We now describe the setting of the online learning algorithm. The buy predictions comes from unbiased experts. The assumption we make is that the buy predictions come from a bounded stochastic distribution such that the loss suffered by the buy experts is i.i.d over rounds. The prediction given to the ski experts is a weighted sum of each buy expert prediction. The ski experts utilize this to recommend a strategy and suffer some loss for the same. The buy expert weights are updated using the Decreasing Hedge algorithm while the weights of the ski experts are updated using the Constant Hedge algorithm. 

\section{Regret Analysis}

In this section, we show our main result - a regret guarantee for the proposed Sequential Ski Rental algorithm.

\begin{theorem}
\label{Claim4}
Let the variance of the best buy expert satisfy $$\gamma_{min} = \frac{\delta \epsilon^{2} }{T c}$$ for some $c \in (1,\infty)$ and the time horizon


\begin{multline*}
    T > \max \{1+\frac{8}{\Delta^{2}}\log \left(\frac{2m}{c-1}\left(1+\frac{Tc\Delta}{\delta \epsilon^{2}}\right)\right), \\ 1+\frac{1}{2\Delta^{2}\log m}\log^{2} \left(\frac{2m}{c-1}\left(1+\frac{Tc\Delta}{\delta \epsilon^{2}}\right)\right),1+\ceil*{\frac{4}{\Delta^{2}}}\} 
\end{multline*}

Then, with probability at least $1-\delta$, the cumulative regret of the Sequential Ski Rental algorithm is bounded  as 

\begin{multline*}
    R_{T} \leq (1+B^{2})\sqrt{T\log n} + 
    B \max \{1+\frac{8}{\Delta^{2}}\log \left(\frac{2m}{c-1}\left(1+\frac{Tc\Delta}{\delta \epsilon^{2}}\right)\right), \\ 1+\frac{1}{2\Delta^{2}\log m}\log^{2} \left(\frac{2m}{c-1}\left(1+\frac{Tc\Delta}{\delta \epsilon^{2}}\right)\right),1+\ceil*{\frac{4}{\Delta^{2}}}\} 
\end{multline*}

where $n$ are the number of ski-experts, where $B$ is the bound on the loss suffered by the ski-experts, $\Delta$ is the minimum sub-optimality gap of the buy experts in terms of their variance, $m \geq 2$ are the number of buy experts and $\epsilon$ is the minimum robustness in terms of the buy cost of the CostRobust Randomized algorithm across rounds.  

\end{theorem}

Recalling the regret definition we use, we have

$$R_{T} = \sum_{t=1}^{T}(\bm{\beta^{t}})^{T}\bm{l}^{t}(b^{t},x^{t},b^{t}_{s},\bm{y^{t}}) - \min_{j}\sum_{t=1}^{T}l^{t}_{j}(b^{t},x^{t},b^{t},y^{t}_{j})$$

This can be split w.r.t the optimal ski expert $j^{*}$, given the true value as $R_{T} = R_{T}^{x} + R_{T}^{b}$ where each component is defined as 

$$R_{T}^{x} = \sum_{t=1}^{T}(\bm{\beta^{t}})^{T}\bm{l}^{t}(b^{t},x^{t},b^{t}_{s},\bm{y^{t}}) - \sum_{t=1}^{T}l^{t}_{j^{*}}(b^{t},x^{t},b^{t}_{s},y^{t}_{j})$$

and 

$$R_{T}^{b} = \sum_{t=1}^{T}l^{t}_{j^{*}}(b^{t},x^{t},b^{t}_{s},y^{t}_{j}) - \sum_{t=1}^{T}l^{t}_{j^{*}}(b^{t},x^{t},b^{t},y^{t}_{j})$$

\begin{theorem}
\label{Claim5}
The first term in the regret split $R_{T}^{x}$ is bounded by
$$R_{T}^{x} \leq (1+B^{2})\sqrt{T\log n}$$
where $B$ is the bound on the loss suffered by the ski-experts and $n$ denotes the number of ski-experts. 
\end{theorem}

This regret bound $R_{T}^{x}$ follows from the standard regret bound for the Constant Hedge algorithm with $n$ experts when losses for each of these experts lie in the range $[0,B]$. To bound $R_{T}^{b}$, note that the loss function is $\epsilon$-robust to the buy cost, hence if the predicted buy cost lies in the range $(b-\epsilon,b+\epsilon)$, this would imply that $l^{t}_{j^{*}}(b^{t},x^{t},b^{t}_{s},y^{t}_{j}) = l^{t}_{j^{*}}(b^{t},x^{t},b^{t},y^{t}_{j})$. Let us analyze the predicted buy cost $b^{t}_{s}$. Note that $\mathbf{E}[b^{t}_{s}|\bm{\alpha^{t}}] = b^{t}$. This is because each of the buy experts in expectation predict correctly and $\sum_{i=1}^{m}\alpha^{t}_{i} = 1$. This leads to $\mathbf{E}[b^{t}_{s}] = b^{t}$. Let $\gamma_{i}$ denote the variance of the $i^{th}$ buy expert. Now to find the variance of $b^{t}_{s}$. 

\begin{equation*}
    \begin{split}
         \mathbf{E}[(b^{t}_{s}-b^{t})^{2}|\bm{\alpha^{t}}] & = \mathbf{E}[(\sum_{i=1}^{m}\alpha^{t}_{i}(a^{t}_{i}-b^{t}))^{2} | \bm{\alpha^{t}}] = \sum_{i=1}^{m}(\alpha^{t}_{i})^{2}\gamma_{i}
    \end{split}
\end{equation*}

where the first equality comes from the fact that $\sum_{i}\alpha^{t}_{i} = 1$, the second equality from the fact that agents predicting at time $t$ are independent and are predicting with mean $b^{t}$. Thus $\mathbf{E}[(b^{t}_{s}-b^{t})^{2}] = \sum_{i=1}^{m}\gamma_{i}\mathbf{E}[(\alpha^{t}_{i})^{2}]$. Using Chebyshev inequality, the probability that $b^{t}_{s}$ lies in the $\epsilon$ range about $b^{t}$ is given as 

$$Pr[|b^{t}_{s}-b^{t}| < \epsilon] > 1 - \frac{var(b^{t}_{s})}{\epsilon^{2}}$$

For this event to hold with probability at least $1-\delta$, we require $\sum_{i=1}^{m}\gamma_{i}\mathbf{E}[(\alpha^{t}_{i})^{2}] \leq \delta \epsilon^{2}$. Under the assumption that one buy expert has variance $\gamma_{min} < \delta \epsilon^{2}$ this is trivially satisfied as $t \to \infty$ .Hence we now require a minimum $t^{*} < T$ such that the above event is satisfied for all $t \in [t^{*},T]$. 

\begin{theorem}
\label{Claim6}
The number of rounds $t^{*}$ after which with probability at least $1-\delta$

$$\sum_{t=t^{*}}^{T}l_{j^{*}}(b^{t},x^{t},b^{t}_{s},y^{t}_{j}) = \sum_{t=t^{*}}^{T}l_{j^{*}}(b^{t},x^{t},b^{t},y^{t}_{j}) ~ \text{is} $$


\begin{multline*}
    t^{*} = \max \{1+\frac{8}{\Delta^{2}}\log \left(\frac{2m}{c-1}\left(1+\frac{Tc\Delta}{\delta \epsilon^{2}}\right)\right), \\ 1+\frac{1}{2\Delta^{2}\log m}\log^{2} \left(\frac{2m}{c-1}\left(1+\frac{Tc\Delta}{\delta \epsilon^{2}}\right)\right),1+\ceil*{\frac{4}{\Delta^{2}}}\}  
\end{multline*}

under the assumption that the variance of one buy expert satisfies $\gamma_{min} = \frac{\delta \epsilon^{2}}{Tc}$ for some $c \in (1,\infty)$, where $m \geq 2$ are the number of buy-experts, $\epsilon$ is the minimum robustness of the CostRobust algorithm in terms of the buy cost across rounds, the sub-optimality gap is $\Delta$ and the time horizon $T > \max \{1+\frac{8}{\Delta^{2}}\log \left(\frac{2m}{c-1}\left(1+\frac{Tc\Delta}{\delta \epsilon^{2}}\right)\right), \\ 1+\frac{1}{2\Delta^{2}\log m}\log^{2} \left(\frac{2m}{c-1}\left(1+\frac{Tc\Delta}{\delta \epsilon^{2}}\right)\right),1+\ceil*{\frac{4}{\Delta^{2}}}\} $.
\end{theorem}

 \begin{proof}
 Consider a time $t_{0} = \ceil*{\frac{4}{\Delta^{2}}}$. If the rate of convergence to the desired variance is upper bounded by $t_{0}$, then we have a convergence rate which does not depend on $T$. If the time taken for convergence is greater than $t_{0}$, then we consider the analysis below. The variance of $b^{t}_{s}$ is 
 $$var(b^{t}_{s}) = \sum_{i=1}^{m}\gamma_{i} \mathbf{E}[(\alpha^{t}_{i})^{2}]$$
 
The update at time $t$ for each buy expert is made based on the squared error, specifically $(a^{t}_{i}-b^{t})^{2}$. We define the loss suffered by the $i^{th}$ buy expert at time $t$ as $g^{t}_{i} = (a^{t}_{i}-b^{t})^{2}$. For every buy expert $i \neq i^{*}$, we define the variable $Z^{t}_{i} := -g^{t}_{i} + g^{t}_{i^{*}} + \Delta_{i}$, which belong to $[-1+\Delta_{i},1+\Delta_{i}]$. We define $G^{t}_{i}$ as the cumulative loss for buy expert $i$ upto time $t$ i.e $G^{t}_{i} = \sum_{s=1}^{t}(a^{s}_{i}-b^{s})^{2}$. Applying Hoeffding's inequality, we get
 
\begin{equation*}
    \begin{split}
        Pr(G^{t-1}_{i}-G^{t-1}_{i^{*}} < \Delta_{i}\frac{t-1}{2}) & = Pr(\sum_{s=1}^{t-1}Z^{s}_{i} > \Delta_{i}\frac{t-1}{2}) \\
     & \leq e^{-(t-1)\frac{\Delta^{2}_{i}}{8}}
    \end{split}
\end{equation*}
 
 When $G^{t-1}_{i}-G^{t-1}_{i^{*}} > \Delta_{i}\frac{t-1}{2}$, then 
 
 \begin{equation*}
 \begin{split}
     \alpha^{t}_{i} & = \frac{e^{-\eta_{b}(G^{t-1}_{i}-G^{t-1}_{i^{*}})}}{1+\sum_{j \neq i^{*}}e^{-\eta_{b}(G^{t-1}_{j}-G^{t-1}_{i^{*}})}} \\
     & \leq e^{-\Delta_{i}\sqrt{(t-1)(\log m) / 2}}
 \end{split}
 \end{equation*}
 
 since $t \geq t_{0} + 1 \geq 2$ .Thus $(\alpha^{t}_{i})^{2} \leq e^{-\Delta_{i}\sqrt{2(t-1)(\log m)}}$. Hence
 
 \begin{equation*}
    \begin{split}
         \mathbf{E}[(\alpha^{t}_{i})^{2}] & \leq Pr(G^{t-1}_{i}-G^{t-1}_{i^{*}} < \Delta_{i}\frac{t-1}{2}) + e^{-\Delta_{i}\sqrt{2(t-1)(\log m)}} \\
         & \leq e^{-(t-1)\frac{\Delta^{2}_{i}}{8}} + e^{-\Delta_{i}\sqrt{2(t-1)(\log m)}} 
    \end{split}
 \end{equation*}
 
 Let us consider each component separately. Considering the first term, since $\Delta_{i} \geq \Delta$ we obtain
 
 \begin{equation*}
     \Delta_{i}e^{-(t-1)\frac{\Delta^{2}_{i}}{8}} \leq \Delta e^{-(t-1)\frac{\Delta^{2}}{8}}
 \end{equation*}
 
 when $ \frac{\Delta\sqrt{t-1}}{2} \geq 1 $ i.e $t \geq 1 + \frac{4}{\Delta^{2}}$ which is satisfied as $t \geq t_{0} + 1 \geq 1 + \frac{4}{\Delta^{2}}$. Now considering the second component
 
 \begin{equation*}
     \Delta_{i}e^{-\Delta_{i}\sqrt{2(t-1)(\log m)}} \leq \Delta e^{-\Delta\sqrt{2(t-1)(\log m)}}
 \end{equation*}
 
 is satisfied if $\Delta\sqrt{2(t-1)(\log m)} \geq 1$ i.e $t \geq 1 + \frac{1}{2 \Delta^{2}\log m}$ which is again ensured by $t \geq t_{0} + 1$.
 
 Also, 
 \begin{equation*}
     e^{-(t-1)\Delta_{i}^{2}/8} \leq e^{-(t-1)\Delta^{2}/8}
 \end{equation*}
 
 and 
 
 \begin{equation*}
     e^{-\Delta_{i}\sqrt{2(t-1)(\log m)}} \leq e^{-\Delta \sqrt{2(t-1)(\log m)}}
 \end{equation*}
 where both these inequalities come from the fact $\Delta_{i} \geq \Delta$.Now, rewriting the variance in terms of the sub-optimality parameters
 
 \begin{equation*}
     \begin{split}
         var(b^{t}_{s}) & = \sum_{i=1}^{m}\gamma_{i}\mathbf{E}[(\alpha^{t}_{i})^{2}] \\
         & \leq \gamma_{min} + \gamma_{min} \sum_{i \neq i^{*}}\mathbf{E}[(\alpha^{t}_{i})^{2}] + \sum_{i \neq i^{*}}\Delta_{i}\mathbf{E}[(\alpha^{t}_{i})^{2}] 
     \end{split}
 \end{equation*}
 
 Hence for every $t \geq t_{0} + 1$, we get
 
 \begin{equation*}
     \begin{split}
         var(b^{t}_{s}) & \leq \gamma_{min} +  m(\gamma_{min} + \Delta)( e^{-\Delta\sqrt{2(t-1)(\log m)}} + e^{-(t-1)\frac{\Delta^{2}}{8}})
     \end{split}
 \end{equation*}
 
 Thus the convergence rate boils down to finding a minimum $t > t_{0}$ such that
 
 \begin{equation*}
     \begin{split}
         e^{-\Delta\sqrt{2(t-1)(\log m)}} +  e^{-(t-1)\frac{\Delta^{2}}{8}} \leq \frac{\frac{\delta \epsilon^{2}}{T} - \gamma_{min}}{m(\gamma_{min} + \Delta)}
     \end{split}
 \end{equation*}
 
Note that we assume $\gamma_{min} = \frac{\delta \epsilon^{2}}{Tc}$ for some $c \in (1,\infty)$. An upper bound on the convergence would be when each of the components is less than $\frac{\delta \epsilon^{2} (c-1)}{2Tm (1 + \rho)\Delta c}$ for some $c \in (1,\infty)$ and $\rho = \frac{\delta \epsilon^{2}}{Tc\Delta}$. Hence after
 

\begin{multline*}
    t^{*} = \max \{1+\frac{8}{\Delta^{2}}\log \left(\frac{2T\Delta mc(1+\rho)}{\delta \epsilon^{2}(c-1)}\right), \\ 1+\frac{1}{2\Delta^{2}\log m}\log^{2} \left(\frac{2T\Delta mc(1+\rho)}{\delta \epsilon^{2}(c-1)}\right)\} 
\end{multline*}

and thus

\begin{multline*}
    t^{*} = \max \{1+\frac{8}{\Delta^{2}}\log \left(\frac{2m}{c-1}\left(1+\frac{Tc\Delta}{\delta \epsilon^{2}}\right)\right), \\ 1+\frac{1}{2\Delta^{2}\log m}\log^{2} \left(\frac{2m}{c-1}\left(1+\frac{Tc\Delta}{\delta \epsilon^{2}}\right)\right)\} 
\end{multline*}


rounds we have that $e^{-\Delta\sqrt{2(t-1)(\log m)}} +  e^{-(t-1)\frac{\Delta^{2}}{8}} \leq \frac{\frac{\delta \epsilon^{2}}{T} - \gamma_{min}}{m(\gamma_{min} + \Delta)}$ for all $t \in [t^{*},T]$. If the above $t^{*} > t_{0}$, then  $var(b^{t}_{s}) < \frac{\delta \epsilon^{2}}{T}$ for all $t \in [t^{*},T]$. If the above $t^{*} \leq t_{0}$, then  $var(b^{t}_{s}) < \frac{\delta \epsilon^{2}}{T}$ for all $t \in [t_{0} + 1,T]$ . Let $t'=\max\{t^{*},t_{0}+1\}$. We would want that for all rounds in $[t',T]$, $b^{t}_{s}$ lies in an $\epsilon$ range around $b^{t}$. Note that we have 

$$Pr[|b^{t}_{s}-b^{t}| < \epsilon] > 1 - \frac{\delta}{T}$$

for all $t \in [t^{'},T]$. We would like to bound the probability $Pr[\bigcap_{t=t'}^{T}\{|b^{t}_{s}-b^{t}| < \epsilon\}]$. Using Frechet inequalities, we have

\begin{equation*}
    \begin{split}
        Pr[\bigcap_{t=t'}^{T}\{|b^{t}_{s}-b^{t}| < \epsilon\}] & \geq \sum_{t=t'}^{T}Pr[|b^{t}_{s}-b^{t}| < \epsilon] - (T-t') \\
        & \geq (1-\frac{\delta}{T})(T-t'+1) - (T-t') \\
        & \geq 1-\frac{\delta}{T} -\frac{\delta}{T}(T-t') \\
        & \geq 1 - \delta
    \end{split}
\end{equation*}

The result follows. 

 
 
 
\end{proof}

\begin{corollary}
\label{Corollary1}
With probability at least $1-\delta$, the second term in the regret split $R_{T}^{b}$ is bounded as 


\begin{multline*}
    R_{T}^{b} \leq B \max \{1+\frac{8}{\Delta^{2}}\log \left(\frac{2m}{c-1}\left(1+\frac{Tc\Delta}{\delta \epsilon^{2}}\right)\right), \\ 1+\frac{1}{2\Delta^{2}\log m}\log^{2} \left(\frac{2m}{c-1}\left(1+\frac{Tc\Delta}{\delta \epsilon^{2}}\right)\right),1+\ceil*{\frac{4}{\Delta^{2}}}\} 
\end{multline*}

where $B$ is the bound on the loss suffered by the ski experts and $\gamma_{min} = \frac{\delta \epsilon^{2}}{Tc}$ for some $c \in (1,\infty)$. 

\end{corollary}

Hence the regret bound follows using the above proved results.

\section{Experiments}

We perform empirical studies to show that our CostRobust algorithm performs similarly to the algorithm proposed by \cite{PSK-18}. We also perform simulations of the \emph{Sequential Ski Rental Algorithm} to verify our theoretical regret guarantees.
\begin{figure*}[ht]
\begin{subfigure}{.24\textwidth}
  \centering
  \includegraphics[width=.99\linewidth]{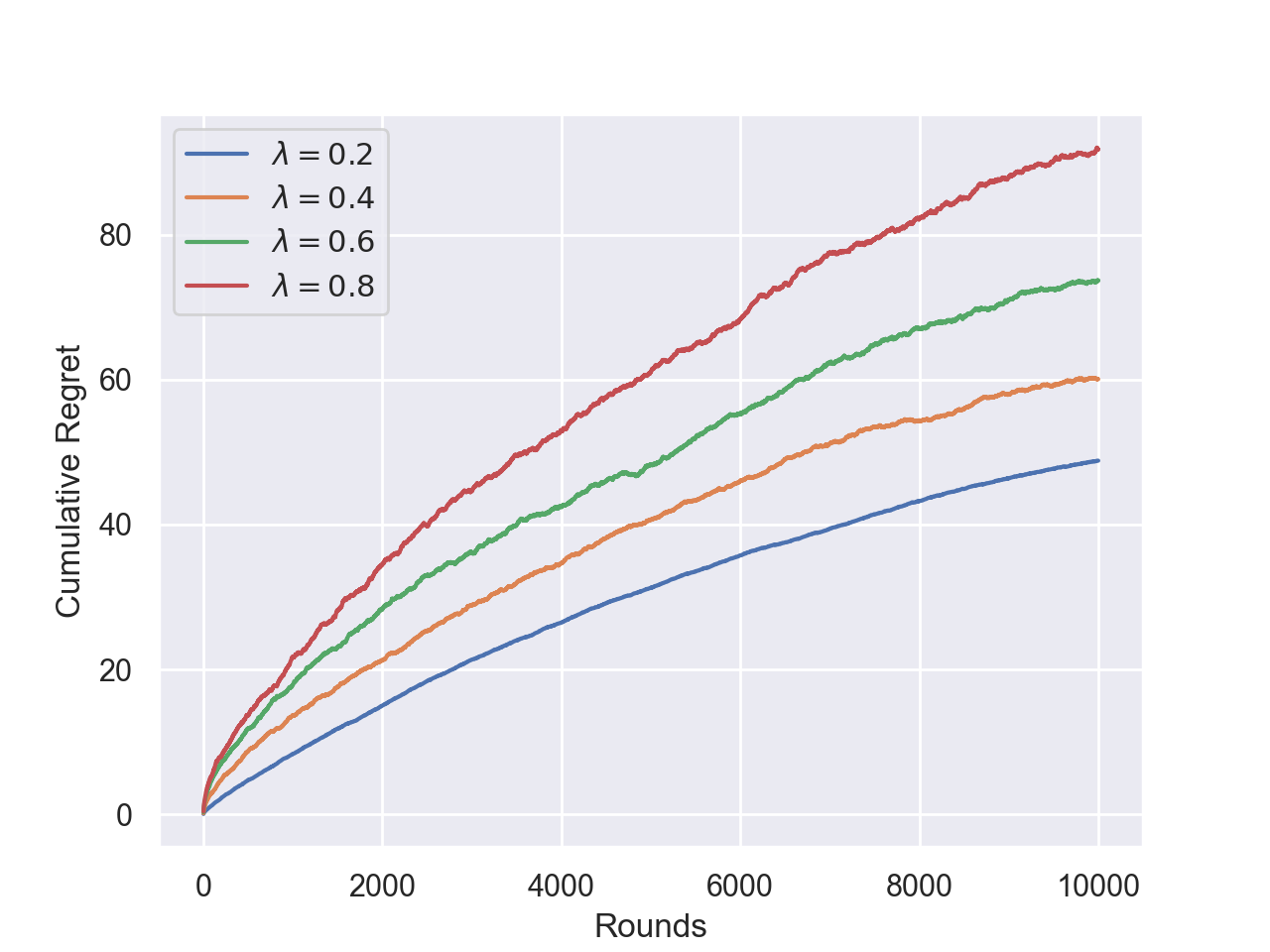}
  \label{fig:lamvar1}
  \caption{$\eta_{min}=1$,$\eta_{max}=100$}
\end{subfigure}%
\begin{subfigure}{.24\textwidth}
  \centering
  \includegraphics[width=.99\linewidth]{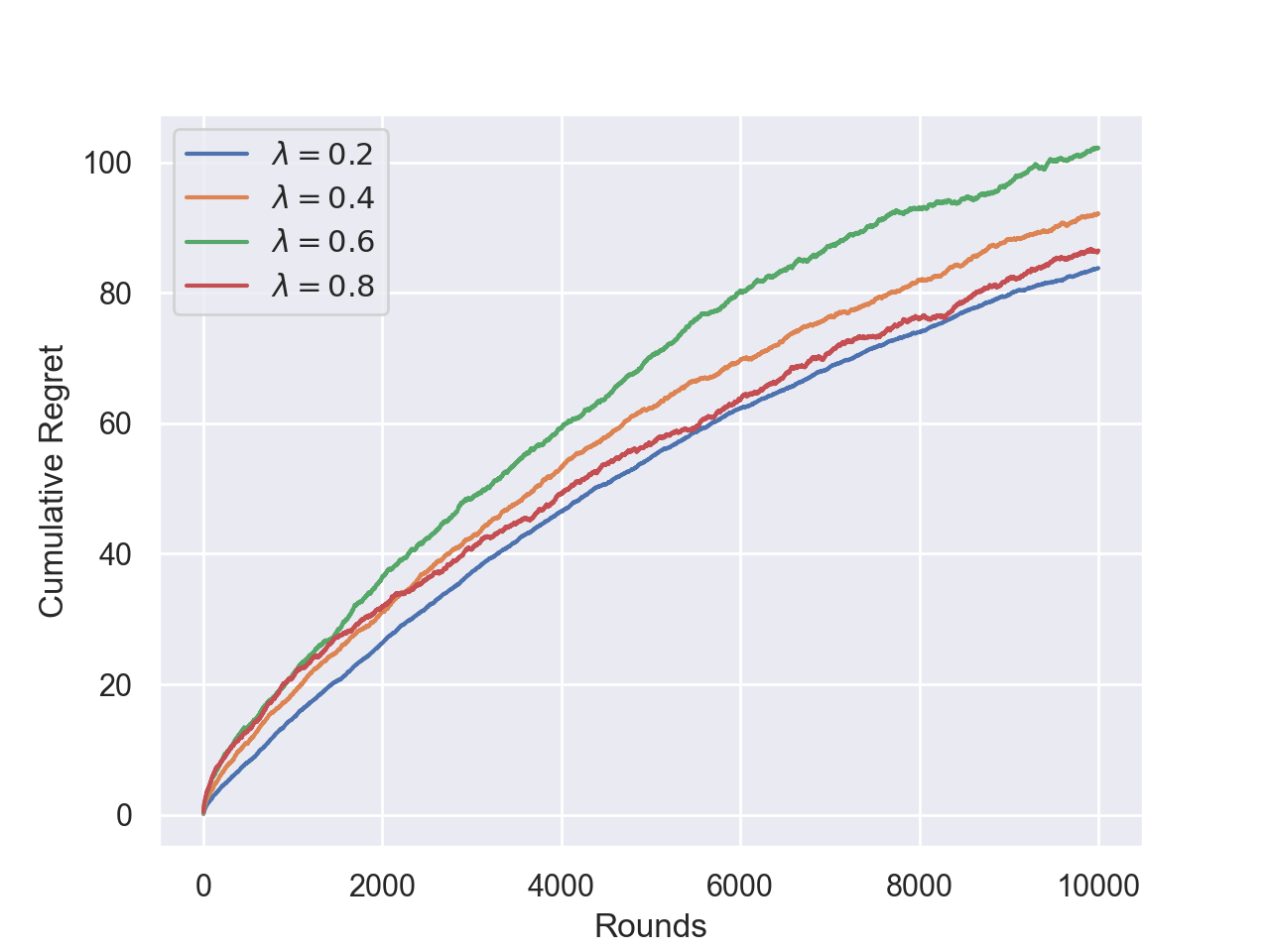}
  \label{fig:lamvar2}
  \caption{$\eta_{min}=100$,$\eta_{max}=150$}
\end{subfigure}
\label{fig:lamvar}
\begin{subfigure}{.24\textwidth}
  \centering
  \includegraphics[width=.99\linewidth]{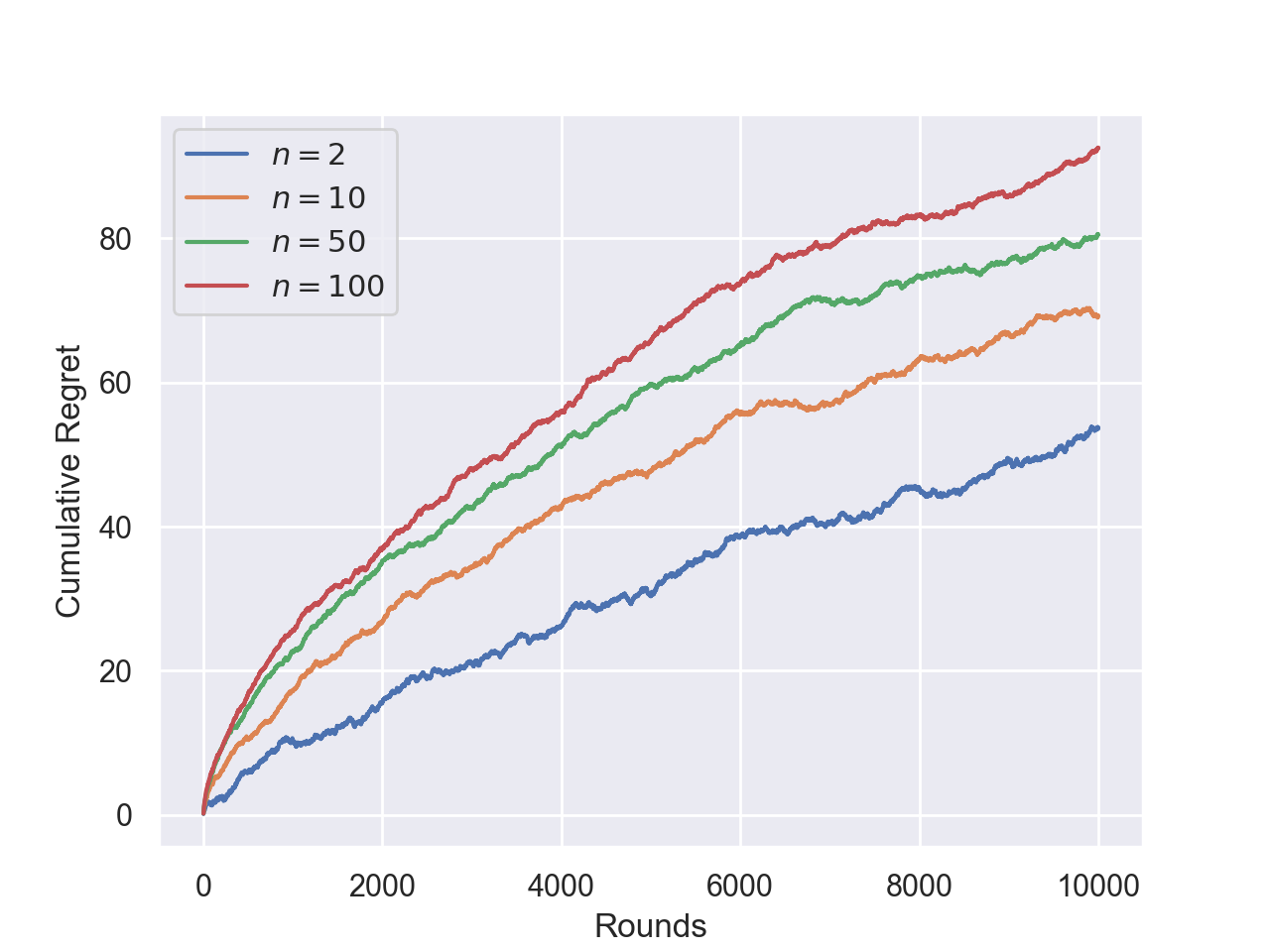}
  \label{fig:svar1}
  \caption{$\eta_{min}=1$,$\eta_{max}=50$}
\end{subfigure}%
\begin{subfigure}{.24\textwidth}
  \centering
  \includegraphics[width=.99\linewidth]{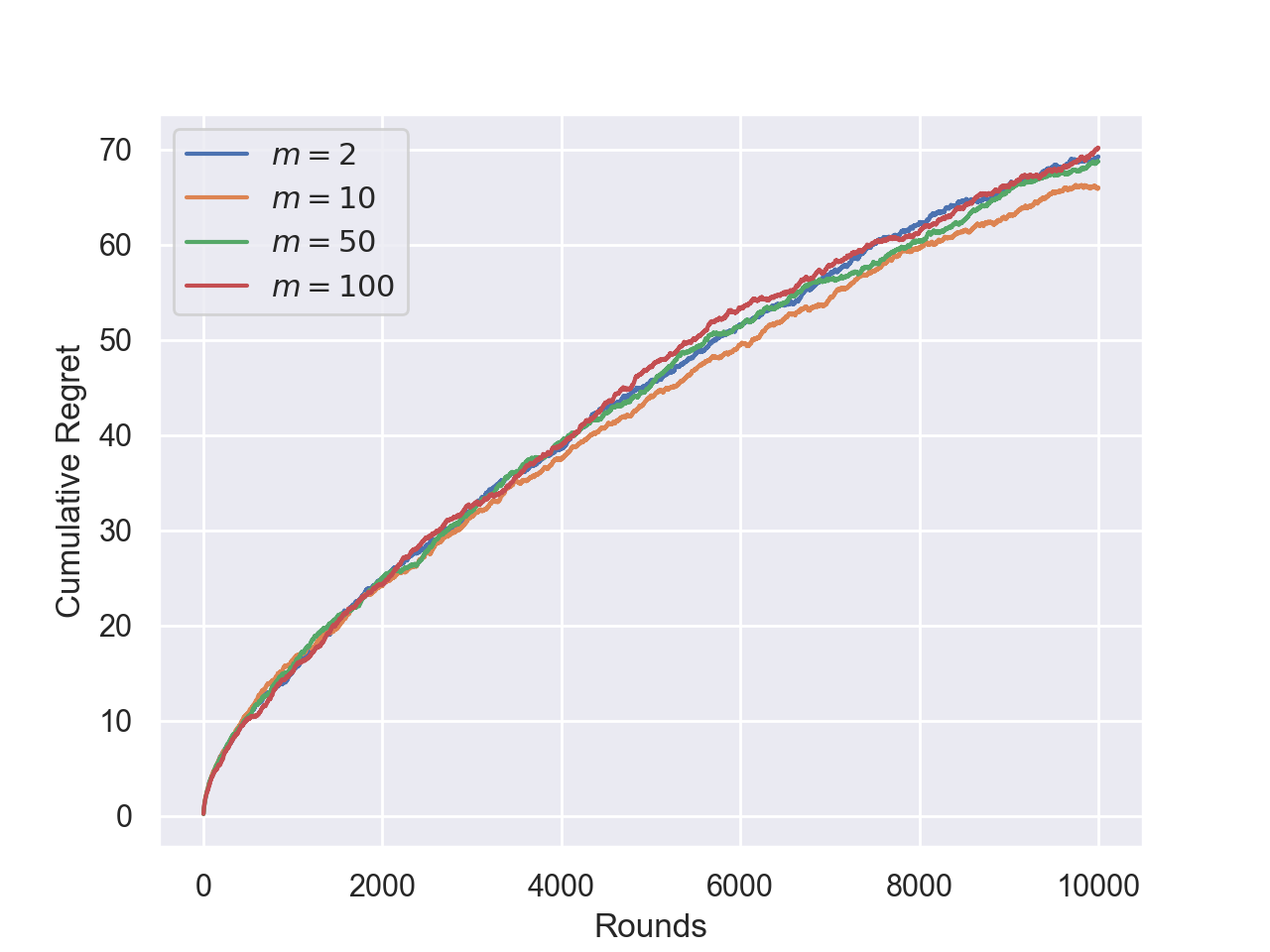}
  \label{fig:svar2}
  \caption{$\eta_{min}=1$, $\eta_{max}=50$}
\end{subfigure}
\caption{((a) and (b)) Regret Variation as a function of the hyper-parameter $\lambda$;
((c) and (d)) Regret Variation as a function of the number of ski experts $n$ and the number of buy experts $m$}
\label{fig:skivar}
\end{figure*}
\subsection{CostRobust Randomized Algorithm}

To show a comparison, we set the cost of buying to $b=100$ and sample the actual number of ski days $x$ uniformly as an integer from $[1,4b]$. To obtain the predicted number of ski-days, we model it as $y = x +\epsilon$, where the noise $\epsilon$ is drawn from a normal distribution with mean $0$ and standard deviation $\sigma$. We compare our algorithms for two different values of the trade-off parameter $\lambda$. For each $\sigma$, we plot the average competitive ratio obtained by each algorithm over 10000 independent trials. 

\begin{figure}
    \centering
    \includegraphics[width=\linewidth]{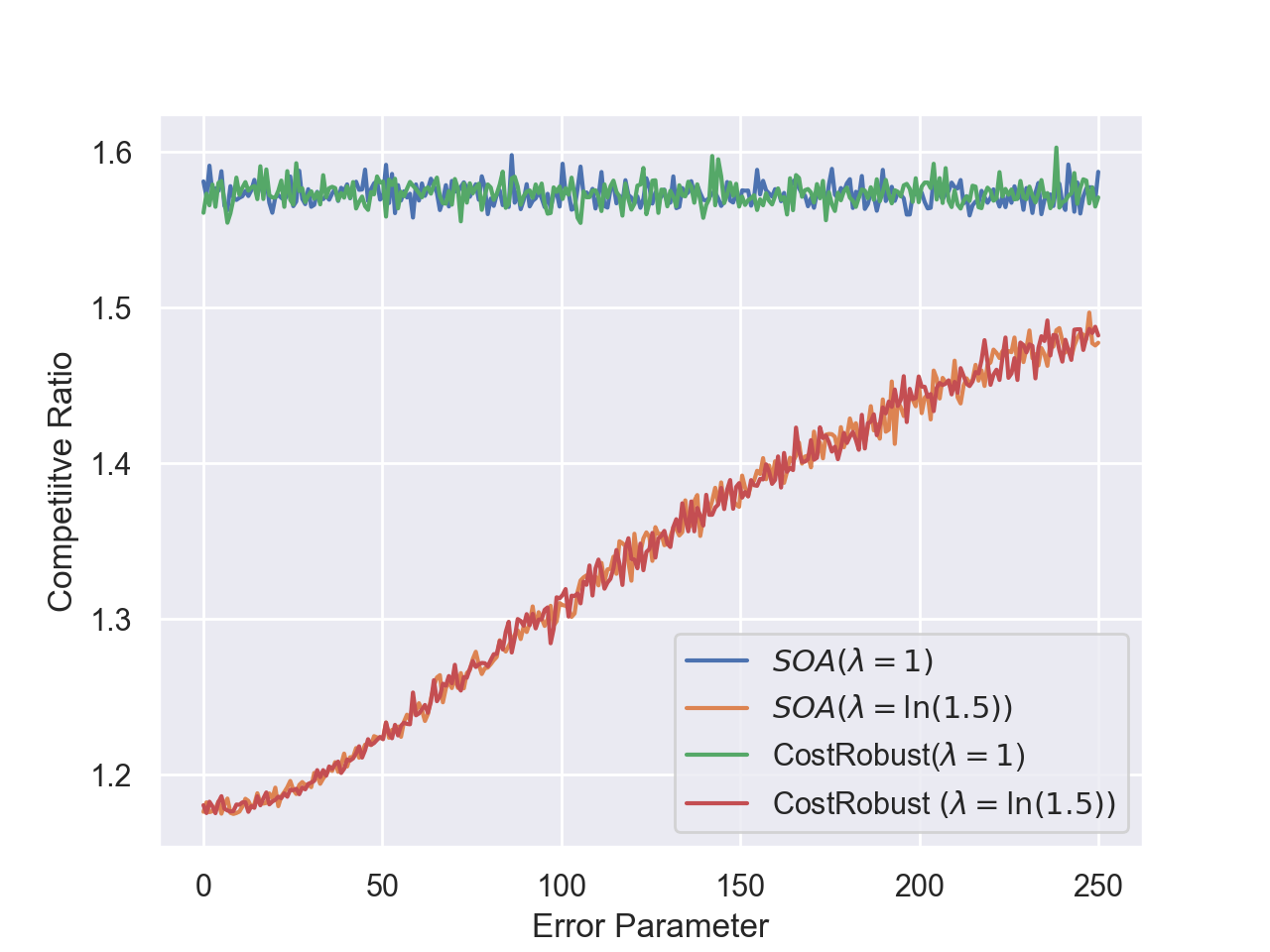}
    \caption{Comparison of the CostRobust randomized algorithm with the randomized algorithm proposed by \cite{PSK-18}}
    \label{fig:compratio}
\end{figure}

Figure \ref{fig:compratio} shows that both the algorithms perform similarly in terms of the competitive ratio. Setting $\lambda=1$, both algorithms ignore the prediction and guarantee a robustness which are close to the theoretical lower bound of $\frac{e}{e-1}$. Setting $\lambda = \ln \frac{3}{2}$ guarantees an upper bound close to $3$ for the CostRobust algorithm. We observe that for such a $\lambda$, the algorithm performs much better than the classical guarantees.

\subsection{Regret Experiments}

While we require certain assumptions to get a theoretical bound, our empirical study shows that we obtain a vanishing regret for much weaker conditions. We obtain regret plots for different settings of $\lambda, n$ and $m$. For all $t \in [1,T]$, $b^{t}$ and $x^{t}$ is a uniformly sampled integer from $[200,700]$. We consider such a range as some experts could have really large errors in predictions which would lead to a negative prediction in the case the bound on the support was lower. We consider $m$ buy-experts and $n$ ski-experts. The learning rate is set according to the Decreasing Hedge algorithm and Constant Hedge algorithm the sets of experts. In our empirical study, the prediction of each buy expert is $a^{t}_{i} = b^{t} + \epsilon_{b}$, where $\epsilon_{b}$ is drawn from a truncated normal distribution in the range $[-50,50]$ with mean $0$ and variance $\gamma_{i}$. For the $m$ buy experts, their variance takes values at uniform intervals from the range $[\gamma_{min},\gamma_{max}]$. Our empirical study uses $\gamma_{min}=1$ and $\gamma_{max}=20$. Note that even though our theoretical bound holds when the noise comes from the range $[-1,1]$ our empirical study shows us that we can achieve vanishing regret for a much weaker constraint. As a modelling choice, we use predictions on the length of the ski season from a normal distribution. The prediction of each ski expert is $y^{t}_{j} = x^{t} + \epsilon_{x}$, where $\epsilon_{x}$ is drawn from a normal distribution with mean $0$ and variance $\eta_{j}$. For the $n$ ski experts, their variance takes values at uniform intervals from the range $[\eta_{min},\eta_{max}]$.The regret plots are obtained over 100 trials. 

\textbf{Variation in $\lambda$} - We expect that if there is a "good" ski expert(a ski expert with less error), then using a lower value of $\lambda$ will give us less algorithmic cost due to the consistency result derived above. However if we do not know the quality of the experts(worst case all of them are bad), the algorithmic cost and hence regret is bounded. The variation is shown in part (a) and (b) Figure \ref{fig:skivar}.
   
  \textbf{Ski Experts Variation} - What the variation in the number of experts shows us is that if we have a few experts at our disposal, making an early decision might be as good as making a decision when the experts have access to the true parameters if not better. An intuition for the learner performing better in the presence of noise is the following situation. Consider the case where $b^{t} < x^{t}$. The optimal strategy is to buy early. If the ski expert predictions are less than $b^{t}$ they would predict sub-optimally when given the true buy cost. If they receive a buy cost sample such that it is less than all of their predictions, then the learner performs better with the noisy sample. However the probability of this decreases as the number of ski-experts increase as all of their predictions need to satisfy this condition. The variation is shown in part (c) of Figure \ref{fig:skivar}. 
    
\textbf{Buy Experts Variation} - The number of buy experts does not affect the cumulative regret as long as the best buy expert comes from a similar error range. This is because the way the learner updates the weights of these experts is based on how far it is from the true buy cost at that time instant. The variation is shown in part (d) of Figure \ref{fig:skivar}.

\section{Conclusion}

In this work, we introduced the sequential ski rental problem, a novel variant of the classical ski buy or rent problem. We developed algorithms and proved regret bounds for the same. Currently we assume that the buy costs are stochastic with different variances. Future work includes considering more general buy cost advice.

\begin{acks}
Arun Rajkumar thanks Robert Bosch Center for Data Science and Artificial Intelligence, Indian Institute of Technology Madras for financial support.
\end{acks}



\bibliographystyle{ACM-Reference-Format} 
\bibliography{sample}


\section*{Appendix}

Below we present the proof of Theorem 3.

\begin{proof}


The hyper-parameter $\lambda \in (\frac{1}{\min_{k \in [T]}b^{k} - 1},1)$.  Consider the case $b^{t} \leq x^{t}$. Thus $OPT=b^{t}$. Note that $d$ is sampled based on two distributions depending on the value of $y^{t}_{j}$ for ski expert $j$. Now in the case $x^{t} \geq d$, we have $ALG = b^{t} + d - 1$, thus for any expert $j$ with this condition

\begin{equation*}
\begin{split}
        l^{t}_{j} & = \frac{d-1}{b^{t}} < \frac{(\min_{k \in [T]}\{b^{k}\}-1)(1+b^{t})}{b^{t}} \\
        & \leq \min_{k \in [T]}\{b^{k}\} - 1 + \frac{\min_{k}\{b^{k}\} - 1}{b^{t}}
    \end{split}
\end{equation*}

where the first inequality comes from the fact that the maximum sampled $d$ can be $\ceil*{\frac{b^{t}_{s}}{\lambda}}$ with $b^{t}_{s} \leq b^{t} + 1$ and $\lambda > 1/(min_{k \in [T]}b^{k} - 1)$. In the case $x^{t} < d$, $ALG = x^{t}$ and thus for any expert $j$ with this condition,

\begin{equation*}
    \begin{split}
        l^{t}_{j} & = \frac{x^{t}-b^{t}}{b^{t}} < \frac{d-b^{t}}{b^{t}} \\
        & < \frac{\min_{k \in [T]}\{b^{k}\}(b^{t}+1) - b^{t}}{b^{t}} \leq \min_{k \in [T]}b^{k}
    \end{split}
\end{equation*}

Consider the case $b^{t} > x^{t}$. Thus $OPT=x^{t}$. In the case $x^{t} < d$ for any ski expert $j$ satisfying that condition, we have $ALG = x^{t}$ and thus $l^{t}_{j} = 0$ for those experts. In the case $x^{t} \geq d$ for any ski expert $j$ in this case, we have $ALG = b^{t} + d - 1$. Thus for these experts,

\begin{equation*}
    \begin{split}
        l^{t}_{j} & = \frac{b^{t} + d - 1 - x^{t}}{x^{t}} < \frac{b^{t}}{x^{t}} \\
        & \leq b^{t}
    \end{split}
\end{equation*}

Thus the worst case bound for any ski expert at time $t$ is $b^{t}$. Since the predictions are unbiased and they are bounded random variables, we have that $b^{t}$ is bounded for each $t$. Thus the loss vectors over $T$ rounds lie in the range $[0,B]$.

\end{proof}

\end{document}